\documentclass[sigconf,nonacm]{acmart}
\usepackage{booktabs}
\usepackage{enumitem}
\usepackage{multirow}
\usepackage{subcaption}

\newcommand{\sstitle}[1]{\vspace{1mm} \noindent {\bf #1}}
\AtBeginDocument{%
  }

\begin{document}

\title{Schema First! Learn Versatile Knowledge Graph Embeddings by Capturing Semantics with MASCHInE}

\author{Nicolas Hubert}
\orcid{0000-0002-4682-422X}
\affiliation{%
 \institution{Université de Lorraine, ERPI, \\
 Université de Lorraine, CNRS, LORIA}
 \city{Nancy}
 \country{France}
}
\email{nicolas.hubert@univ-lorraine.fr}

\author{Heiko Paulheim}
\orcid{0000-0003-4386-8195}
\affiliation{%
 \institution{University of Mannheim}
 \city{Mannheim}
 \country{Germany}}
\email{heiko.paulheim@uni-mannheim.de}

\author{Pierre Monnin}
\orcid{0000-0002-2017-8426}
\affiliation{%
 \institution{Univ. Côte d’Azur, Inria, CNRS, I3S}
 \city{Sophia Antipolis}
 \country{France}
}
\email{pierre.monnin@inria.fr}

\author{Armelle Brun}
\orcid{0000-0002-9876-6906}
\affiliation{%
\institution{Université de Lorraine, CNRS, LORIA}
\city{Nancy}
\country{France}}
\email{armelle.brun@loria.fr}

\author{Davy Monticolo}
\orcid{0000-0002-4244-684X}
\affiliation{%
 \institution{Université de Lorraine, ERPI}
 \city{Nancy}
 \country{France}}
\email{davy.monticolo@univ-lorraine.fr}

\renewcommand{\shortauthors}{Hubert et al.}

\begin{abstract}
Knowledge graph embedding models (KGEMs) have gained considerable traction in recent years. These models learn a vector representation of knowledge graph entities and relations, a.k.a. knowledge graph embeddings (KGEs). Learning versatile KGEs is desirable as it makes them useful for a broad range of tasks. However, KGEMs are usually trained for a specific task, which makes their embeddings task-dependent.
In parallel, the widespread assumption that KGEMs actually create a semantic representation of the underlying entities and relations (\textit{e.g.}, project similar entities closer than dissimilar ones) has been challenged. In this work, we design heuristics for generating \emph{protographs} -- small, modified versions of a KG that leverage RDF/S information. The learnt protograph-based embeddings are meant to encapsulate the semantics of a KG, and can be leveraged in learning KGEs that, in turn, also better capture semantics.
Extensive experiments on various evaluation benchmarks demonstrate the soundness of this approach, which we call \textbf{M}odular and \textbf{A}gnostic \textbf{SCH}ema-based \textbf{In}tegration of protograph \textbf{E}mbeddings (MASCHInE). In particular, MASCHInE helps produce more versatile KGEs that yield substantially better performance for entity clustering and node classification tasks. For link prediction, using MASCHinE substantially increases the number of semantically valid predictions with equivalent rank-based performance.
\end{abstract}




\maketitle
\section{Introduction}
\label{introduction}
Knowledge Graphs (KGs) such as DBpedia and YAGO represent facts as triples $(h,r,t)$ formed by a head $h$ and a tail $t$ linked by a semantic relation $r$ which qualifies the nature of their relationship. 
Typical learning problems with KGs are Link Prediction (LP), Entity Clustering (EC), and Node Classification (NC).

In most cases, these problems are addressed using
Knowledge graph embedding models (KGEMs),
which generate vector representations for entities and relations of a KG, a.k.a. Knowledge Graph Embeddings (KGEs). While generating dense and numerical vectors for entities and relations, KGEMs are expected to produce KGEs that retain the underlying semantics of the KG~\cite{wang-2021}. 
This widespread assumption that KGEMs create a semantic representation of the underlying entities and relations (i.e., project similar entities closer than dissimilar ones) has been challenged recently~\cite{jain-2021}. Their semantic capabilities presumably inherited from the rich information contained in the KG is neither fully satisfying nor consistent for tasks such as concept clustering~\cite{jain-2021}. In this work, we claim that this is because schemas (\textit{e.g.} RDF/S and OWL) are often overlooked as an interesting source of information for improving embeddings.

Injecting schema-based information into the training phase could help enhancing the semantic awareness of KGEMs~\cite{hubert-2023-loss}. This way, better performance can be reached using available ontological information, without introducing major overhead in models' complexity. A few works incorporate such schemas when training KGEM, whether it is in the loss function~\cite{damato-2021, hubert-2023-loss}, in the negative sampling procedure~\cite{krompass-2015, jain-iswc}, or in the model representations~\cite{tkrl, tarp}. All of those methods are usually bound to just one KGEM and therefore not universally applicable.

Another line of research that is less explored focuses on preprocessing graphs prior to encoding their components~\cite{chen-2018, pietrasik-2023}, rather than focusing on the representation capability of the KGEM itself.
This idea of summarizing~\cite{cebiric-2019, goasdoue-2020} or coarsening~\cite{pietrasik-2023} RDF\footnote{Resource Description Framework. See \url{https://www.w3.org/RDF}} graphs proves useful in many circumstances. The intended goals and expected benefits are multiple: scaling up training~\cite{huang-2021}, initializing embeddings with more robust vector representations learnt on the coarsened graph~\cite{pietrasik-2023}, and improving predictive performance in KG-based downstream tasks~\cite{wang-2022-kdd}. It is worth a reminder that several heuristics exist for coarsening or summarizing graphs~\cite{cebiric-2019}.
All these approaches, however, use statistics over the graphs, instead of considering schema information. In contrast, our work focuses on using RDF/S information to generate an abstraction of the KG that encapsulates general knowledge about how entities and relations interact based on entity types, and relation domains and ranges. We call such an abstraction a \emph{protograph} (see left part of Figure~\ref{fig:pipeline}).

If considering schema-based information is usually done with the goal of improving results w.r.t. a given task such as LP~\cite{damato-2021, hubert-2023-loss}, it would be worth investigating whether the learnt embeddings actually have stronger semantic capabilities~\cite{jain-2021, hubert-2023-loss}. If so, they are expected to provide better performance with regards to other tasks such as entity classification and clustering.
In this work, we propose an approach named MASCHInE that leverages RDF/S information in a model-agnostic way, making it applicable to arbitrary KGEMs. Specifically, we create and embed \emph{protographs} from a KG schema. The protograph embeddings are then used to initialize the actual KGEs, and training starts on these pre-trained KGEs. Finally, we assess the versatility of these KGEs by comparing results with a vanilla, traditional learning approach.

The main contributions of our work are summarized as follows:
\begin{itemize}
    \item We devise two heuristics for building protographs that aim at encapsulating concise and meaningful information derived from the schema that underpins a KG.
    \item To the best of our knowledge, we propose the first work that studies the potential of pre-training embeddings with a protograph-assisted approach based on a schema, and subsequently use these embeddings for multiple tasks, \textit{e.g.} link prediction, entity clustering and node classification.
    \item Through extensive experiments on several benchmarks, we demonstrate the value of using schema-based protographs as a means to generate versatile, multi-purpose embeddings.
\end{itemize}

The remainder of the paper is structured as follows. 
Related work is presented in Section~\ref{related-work}. 
In Section~\ref{approach}, we detail our approach for building protographs and how they fit into our approach MASCHInE. In Section~\ref{experiments}, the potential usefulness of embeddings learnt with MASCHInE is discussed, and experiments w.r.t. several benchmarks are carried out.
Lastly, Section~\ref{conclusion} sums up the main findings and outlines promising future research directions.

\begin{figure*}[t]
  \centering
  \includegraphics[scale=0.55]{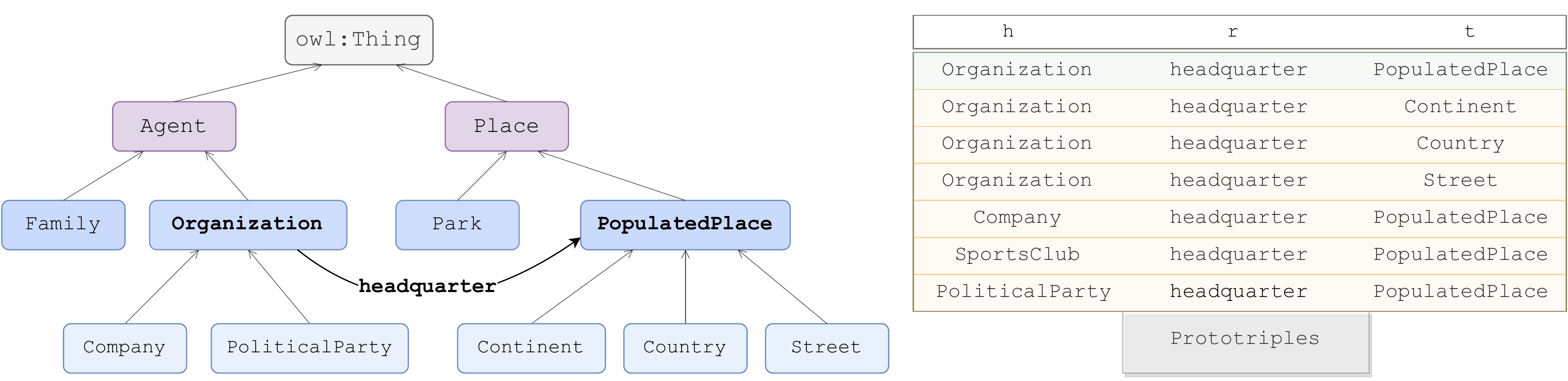}
  \caption{Excerpt from DBpedia class hierarchy (left) and the generated prototriples (right) according to P2. The full URI names have been shortened due to space limitations. The prototriple highlighted in green (right) comes from the schema, whereas the following triples are generated by fixing the domain (resp. range) of \texttt{dbo:headquarter} and replacing the class in the range (resp. domain) with each of its direct subclasses.}
  \label{fig:protograph2}
\end{figure*}

\section{Related Work}
\label{related-work}
\sstitle{Knowledge graph embeddings.} KGEMs have gained significant attention in recent years due to their ability to represent structured knowledge in a continuous vector space. The seminal translational model TransE~\cite{transe} represents entities and relations as low-dimensional vectors and defines the relationship between a head, a relation, and a tail using a translation operation in the embedding space. Most of the KGEMs subsequently proposed aim at addressing its limitations and improve the representation expressiveness of KGEs~\cite{wang-2021}, such as DistMult~\cite{distmult}, ComplEx~\cite{complex}, ConvE~\cite{conve}, and TuckER~\cite{tucker}. Embeddings learnt by these KGEMs demonstrated potential applicability in tasks such as LP~\cite{wang-2021}.

\sstitle{Schema-based representation learning.} 
Many knowledge graphs are backed by a more or less expressive schema defining which classes of entities exist and by which relations they are allowed to interact, among others. Recent approaches in KG-based learning leverage the schema as a supplementary source of information to learn higher-quality embeddings for the task at hand. A common approach is to alter the negative sampling procedure in order to produce more realistic negative triples. For instance, type-constrained negative sampling (TCNS)~\cite{krompass-2015} replaces the head or the tail with a random entity belonging to the same type as the ground-truth entity.
The model itself can be adapted to take schema-based information into account. In TaRP~\citep{tarp}, type and entity-level information are simultaneously considered and encoded as prior probabilities and likelihoods of relations, respectively.
Altering the loss function with schema-based information is another line of research. In~\citep{damato-2021}, background ontological information is injected in the pairwise hinge loss function as constraints. Hubert \textit{et al.}~\citep{hubert-2023-loss} leverage domains and ranges of relations, and propose semantic-driven loss functions that extend the most frequently used loss functions in LP.

\section{Proposed Approach}
In this section, we detail the two heuristics we propose for building protographs using schema-based information. Next, we provide some perspective on the overall proposed approach MASCHInE.
\label{approach}
\subsection{Protographs}
\label{protographs}
Most KGs are underpinned by a schema, \textit{e.g.}, DBpedia, Wikidata, and YAGO. Generating a protograph using schema-based information to improve KGEM performance with respect to a given task is an under-explored avenue.

The protographs we build leverage RDF/S information and contain an entity $p_C$ for each class $C$, and the same set of relations as the KG. The first design principle consists in adding an edge for each relation with domain and range restrictions, \textit{i.e.}, given the two axioms $domain(r,C_i)$, $range(r,C_j)$, we add a triple $(p_{C_i},r,p_{C_j})$ to the protograph\footnote{Note that $C_i$ and $C_j$ might be the same, so there can be cycles in the protograph.}. We refer to this strategy as \textbf{P1}.

The protograph P1 contains as many triples as there are relations with both a domain and range\footnote{In the KGs used in this paper, all relations have an explicit domain and range.} in the KG.
This may lead to relatively small-sized protographs from which there is little data for the KGEM to learn from. In addition, relation domains and ranges can refer to generic classes.
If entities are typed in a fine-grained way -- \textit{i.e.} with more specific classes -- such classes will be absent from the protograph. As such, we propose a second design principle. Assuming we observe the following axioms $domain(r,C_i)$, $range(r,C_j)$, $subclassOf(C_i',C_i)$, $subclassOf(C_j',C_j)$, we add the following triples to the protograph: $(p_{C_i},r,p_{C_j})$, $(p_{C_i'},r,p_{C_j})$, $(p_{C_i}$, $r,p_{C_j'})$.
In other words: for each subclass of the class appearing in the domain or range of a relation, an additional triple is created. 

We refer to this protograph as \textbf{P2}. Figure~\ref{fig:protograph2} illustrates this process for a triple extracted from DBpedia. 

P2 usually contains more triples than P1, at the possible expense of containing dubious ones. To illustrate, the DBpedia ontology contains the triples: $domain(currentWorldChampion,Sport)$, $range(currentWorldChampion$, $Agent)$, and $subclassOf($$Family$, $Agent)$. They lead to the triple $(P_{Sport},currentWorldChampion$, $P_{Family})$ for which the relevancy and correctness are questionable. 
Conceptually, with P2, we accept some noise to be introduced in the protograph for the benefit of more triples being generated. 
However, the generation of dubious triples is limited by two factors: firstly, when adding triples that derive from an original triple (\textit{i.e.} containing the domain and range of the relation), we only consider direct subclasses -- in contrast to transitive subclasses. Secondly, we fix one side of the triple and generate as many new triples as there are direct subclasses of the class of the other side, i.e., we avoid generating triples $(p_{C_i'},r,p_{C_j'})$.

\begin{figure*}[t]
  \centering
  \includegraphics[scale=0.595]{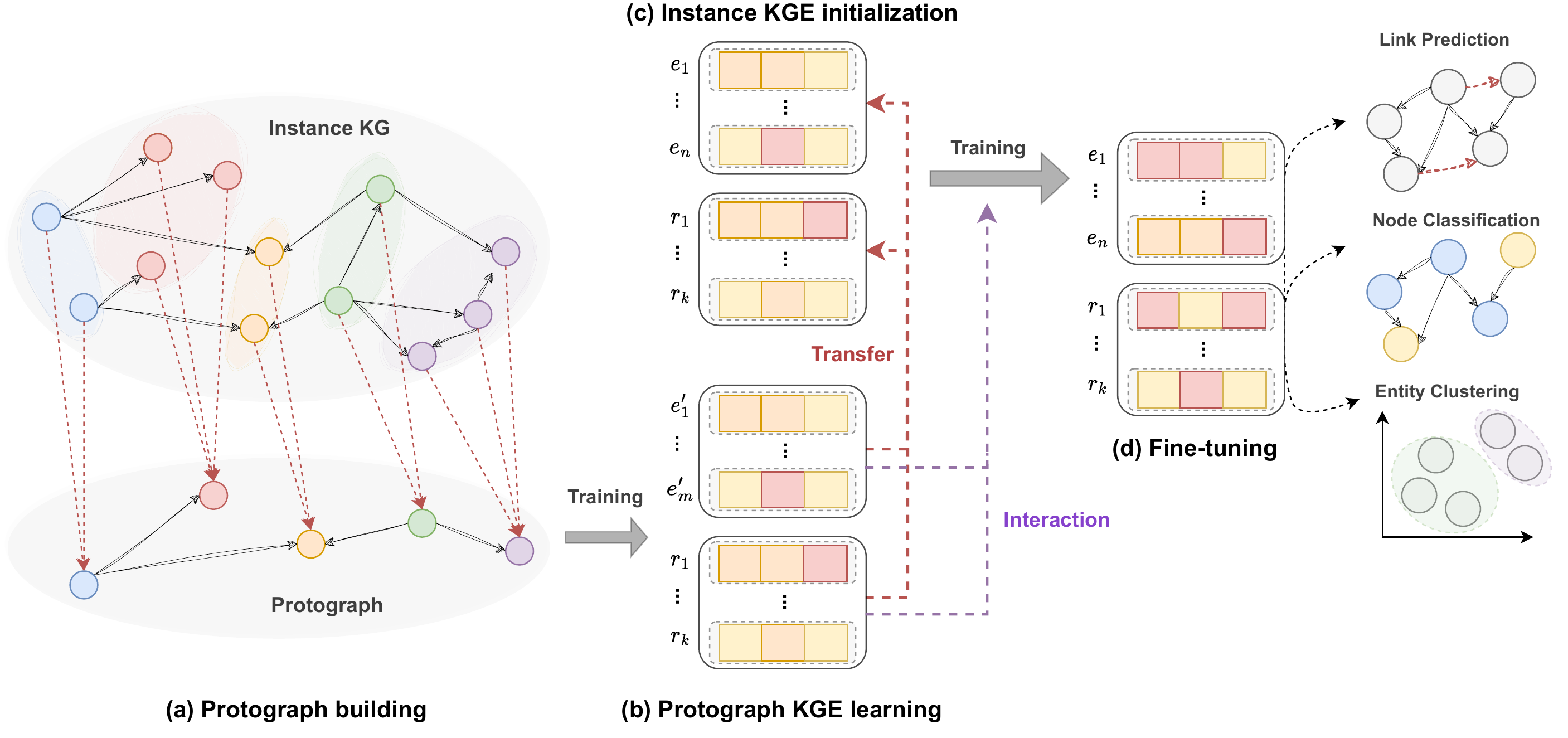}
  \caption{Overview of MASCHInE. The overall training approach is based on several modules that each features different possibilities. First, the protograph is built following predefined design principles (a). Protograph embeddings are learnt (b) before being transferred to the corresponding entities and relations of the KG (c). Next, training starts on the KG (d). During this step, embeddings learnt on the protograph can still interact in the fine-tuning procedure (purple dotted arrow). Finally, the fine-tuned KGEs can be used in several tasks.}
  \label{fig:pipeline}
\end{figure*}

\subsection{MASCHInE}
\label{framework}
The proposed MASCHInE approach allows for generating pre-trained embeddings based on a protograph, to be further fine-tuned on the initial KG. MASCHInE is intended to produce more versatile KGEs that can be successfully used in various tasks.
Conceptually, MASCHInE is made of four different modules that are depicted in Figure~\ref{fig:pipeline} and further detailed below.

\textbf{(a) Protograph building} is the preliminary step that generates a second source of data to learn from. In particular, schema-based information is leveraged to create a protograph that encapsulates more general knowledge about how entities and relations interact based on ontological constraints such as entity types, relation domains and ranges, etc. A mapping dictionary is created to pair entities between the KG and the protograph (the set of relations in the KG and the protograph are identical). In this work, each KG entity maps to its (potentially multiple) most specific classes -- represented as entities in the protograph. In Section~\ref{protographs}, we presented two heuristics for building protographs with such constraints. Other possibilities exist and we leave them for future work.

\textbf{(b) Protograph KGE learning} is the task of learning KGEs on the protograph itself. This can be achieved regardless of the KGEM at hand and is therefore model-agnostic\footnote{In our experiments, the same KGEM is used for both pre-training over the protograph and fine-tuning embeddings on the KG.}.

\textbf{(c) Instance KGE initialization} consists in transferring protograph embeddings to the corresponding entity and relation embeddings of the KG. To do so, we assign each entity in the KG the embedding vector of its corresponding protograph entity using the dictionary created in step (a) (\textit{i.e.}, its most specific class). When an entity maps to multiple protograph entities (\textit{i.e.} when there exist mutiple most specific classes for a given entity), the protograph embeddings of its most specific classes are averaged and the resulting embedding is transferred to the entity.

\textbf{(d) Fine-tuning} is performed on the entity KGEs. At this stage, embeddings learnt on the protograph could still intervene in the fine-tuning procedure. In this work, however, we stick to plain entity KGE initialization, \textit{i.e.} protograph embeddings are frozen after being transferred, and do not interact with entity KGEs anymore.

\section{Experiments}
\label{experiments}
In our experiments, we apply the MASCHInE approach presented in Section~\ref{approach}.
We experiment with five popular KGEMs. We first train on protographs, and then use the protograph embeddings to initialize entity and relation embeddings in the KG. Training starts on the KG and the fine-tuned KGEs are ultimately assessed in three tasks, namely LP, EC, and NC. For each task, we compare the vanilla setting (V), \textit{i.e.}, directly training the respective KGEM on the KG, with the MASCHInE approach based on P1 (resp. P2). Datasets, optimal hyperparameters, scripts, and pre-trained embeddings are publicly released\footnote{\href{https://github.com/nicolas-hbt/versatile-embeddings}{https://github.com/nicolas-hbt/versatile-embeddings}}.
In the following, protographs, KG entities and relations are encoded using five mainstream KGEMs: TransE~\cite{transe}, DistMult~\cite{distmult}, ComplEx~\cite{complex}, ConvE~\cite{conve}, and TuckER~\cite{tucker}.

\subsection{Link Prediction}
\label{LP}

\begin{table}[h]
\caption{KG and protograph characteristics. Column headers from left to right: number of entities, relations, and triples.}\label{tab:datasets}
\centering
\footnotesize
\begin{tabular}{lrrrr}
\toprule 
Dataset & & $|\mathcal{E}|$ & $|\mathcal{R}|$ & $|\mathcal{T}|$\\
\midrule
\multirow{3}{*}{YAGO14k}
& KG & $14,178$ & $37$ & $19,183$ \\
& P1 & $22$ & $37$ & $37$ \\
& P2 & $590$ & $37$ & $4,959$ \\
\midrule
\multirow{3}{*}{FB15k187}
& KG & $14,305$ & $187$ & $278,436$ \\
& P1 & $138$ & $187$ & $187$ \\
& P2 & $138$ & $187$ & $187$ \\
\midrule
\multirow{3}{*}{DBpedia77k}
& KG & $76,651$ & $150$ & $190,028$ \\
& P1 & $55$ & $150$ & $150$ \\
& P2 & $186$ & $150$ & $3,210$ \\
\bottomrule
\end{tabular}
\end{table}

\begin{table*}[t]
\centering
\caption{Link prediction results on YAGO14k, FB15k187, and DBpedia77k using KGEMs trained without protograph (V), with P1, and P2. For a given dataset and KGEM, comparisons are made between the three settings, and best results are in bold fonts. Hits@K and Sem@K are abbreviated as H@K and S@K, respectively.}\label{tab:LP-results}
\footnotesize
\begin{tabular}{ccccccccccccccccccc}
\toprule
& & \multicolumn{5}{c}{YAGO14k} && \multicolumn{5}{c}{FB15k187} && \multicolumn{5}{c}{DBpedia77k} \\
\cmidrule(lr){3-7} \cmidrule(lr){9-13} \cmidrule(lr){15-19}
& & MRR & H@3 & H@10 & S@3 & S@10 & & MRR & H@3 & H@10 & S@3 & S@10 & & MRR & H@3 & H@10 & S@3 & S@10\\
\midrule
\multirow{3}{*}{TransE} & V & $\mathbf{0.766}$ & $\mathbf{0.806}$ & $\mathbf{0.885}$ & $0.987$ & $0.971$ & & $\mathbf{0.232}$ & $\mathbf{0.259}$ & $\mathbf{0.430}$ & $\mathbf{0.970}$ & $0.958$ 
&& $\mathbf{0.236}$ & $\mathbf{0.285}$ & $\mathbf{0.405}$ & $0.983$ & $0.976$\\
& P1 & $0.746$&$0.791$&$0.871$&$0.994$&$0.993$& &$0.227$&$0.254$&$0.425$ & $\mathbf{0.970}$ & $\mathbf{0.959}$& &$0.217$&$0.260$&$0.382$&$0.985$&$0.980$\\
& P2 & $0.700$ & $0.762$ & $0.854$ & $\mathbf{1.000}$ & $\mathbf{0.999}$ && $0.227$ & $0.254$ & $0.425$ & $\mathbf{0.970}$ & $\mathbf{0.959}$ 
&& $0.216$ & $0.260$ & $0.380$ & $\mathbf{0.987}$ & $\mathbf{0.982}$ \\
\midrule
\multirow{3}{*}{DistMult} & V & $\mathbf{0.899}$ & $\mathbf{0.916}$ & $\mathbf{0.926}$ & $0.658$ & $0.490$ 
&& $\mathbf{0.228}$ & $\mathbf{0.256}$ & $\mathbf{0.412}$ & $0.985$ & $0.969$ && $\mathbf{0.282}$ & $\mathbf{0.308}$ & $\mathbf{0.391}$ & $0.845$ & $0.804$ \\
& P1 & $0.164$ & $0.169$ & $0.314$ & $\mathbf{1.000}$ & $\mathbf{1.000}$ && $0.206$ & $0.228$ & $0.394$ & $\mathbf{0.988}$ & $\mathbf{0.980}$ 
&& $0.194$ & $0.219$ & $0.303$ & $0.928$ & $0.939$ \\
& P2 & $0.289$ & $0.325$ & $0.511$ & $0.999$ & $0.999$ && $0.206$ & $0.228$ & $0.394$ & $\mathbf{0.988}$ & $\mathbf{0.980}$ 
&& $0.190$ & $0.207$ & $0.287$ & $\mathbf{0.937}$ & $\mathbf{0.947}$ \\
\midrule
\multirow{3}{*}{ComplEx} & V & $\mathbf{0.923}$ & $\mathbf{0.930}$ & $0.933$ & $0.723$ & $0.587$ && $\mathbf{0.203}$ & $\mathbf{0.223}$ & $\mathbf{0.399}$ & $\mathbf{0.942}$ & $\mathbf{0.923}$ 
&& $\mathbf{0.286}$ & $\mathbf{0.317}$ & $\mathbf{0.394}$ & $0.823$ & $0.754$ \\
& P1 & $0.897$ & $0.916$ & $0.932$ & $0.907$ & $0.848$ && $0.185$ & $0.201$ & $0.371$ & $0.922$ & $0.900$ && $0.209$ & $0.226$ & $0.302$ & $0.856$ & $0.839$ \\
& P2 & $0.914$ & $\mathbf{0.930}$ & $\mathbf{0.934}$ & $\mathbf{0.914}$ & $\mathbf{0.854}$ && $0.185$ & $0.201$ & $0.371$ & $0.922$ & $0.900$ && $0.209$ & $0.225$ & $0.306$ & $\mathbf{0.913}$ & $\mathbf{0.890}$ \\
\midrule
\multirow{3}{*}{ConvE} & V & $\mathbf{0.934}$ & $\mathbf{0.936}$ & $\mathbf{0.940}$ & $0.858$ & $0.824$ && $0.265$ & $0.297$ & $0.463$ & $\mathbf{0.979}$ & $\mathbf{0.971}$ && $0.227$ & $\mathbf{0.259}$ & $0.352$ & $0.897$ & $0.889$ \\
& P1 & $0.927$ & $0.930$ & $0.935$ & $0.882$ & $0.845$ && $0.272$ & $0.301$ & $\mathbf{0.470}$ & $0.977$ & $0.970$ && $0.227$ & $0.256$ & $0.352$ & $0.932$ & $0.932$ \\
& P2 & $0.931$ & $0.931$ & $0.938$ & $\mathbf{0.935}$ & $\mathbf{0.922}$ && $\mathbf{0.272}$ & $\mathbf{0.301}$ & $\mathbf{0.470}$ & $0.977$ & $\mathbf{0.971}$ && $\mathbf{0.229}$ & $\mathbf{0.259}$ & $\mathbf{0.356}$ & $\mathbf{0.943}$ & $\mathbf{0.942}$ \\
\midrule
\multirow{3}{*}{TuckER} & V & ${0.919}$ & ${0.931}$ & $\mathbf{0.942}$ & $0.899$ & $0.821$ && $0.282$ & $0.311$ & $0.467$ & $\mathbf{0.996}$ & $\mathbf{0.991}$ && $0.274$ & ${0.305}$ & $0.401$ & ${0.988}$ & ${0.985}$ \\
& P1 & $\mathbf{0.922}$ & $\mathbf{0.933}$ & $0.941$ & $0.951$ & $0.925$ && $\mathbf{0.286}$ & $\mathbf{0.317}$ & $\mathbf{0.472}$ & $0.995$ & $0.985$ && $\mathbf{0.277}$ & $\mathbf{0.308}$ & $0.402$ & ${0.987}$ & ${0.984}$ \\
& P2 & $0.918$ & $0.930$ & $0.941$ & $\mathbf{0.971}$ & $\mathbf{0.939}$ && $\mathbf{0.286}$ & $\mathbf{0.317}$ & $\mathbf{0.472}$ & $0.995$ & $0.985$ && ${0.275}$ & $0.304$ & $\mathbf{0.403}$ & $\mathbf{0.989}$ & $\mathbf{0.986}$ \\
\bottomrule
\end{tabular}
\end{table*}

\sstitle{Datasets.} Due to the need for schema-defined KGs, the following three benchmark datasets are considered: YAGO14k, FB15k187, and DBpedia77k~\cite{hubert-2023-loss}.
In particular, all entities are typed and relations have both a defined domain and range. Table~\ref{tab:datasets} provides statistics for these datasets and their respective protographs P1 and P2. Note that protograph entities are classes from the original KG. Full dataset description is provided at: \url{https://github.com/nicolas-hbt/versatile-embeddings\#datasets}.

\sstitle{Evaluation metrics.} This section is motivated by evaluating the fine-tuned KGEs for LP. Performance is assessed w.r.t. Mean Reciprocal Rank (MRR), Hits@$K$, and Sem@$K$~\cite{hubert-2022,hubert-2023-sem}. The latter accounts for the proportion of
triples whose predicted head (resp. tail) belongs to the domain (resp. range) of the relation.

\sstitle{Implementation details.} KGEMs were implemented in PyTorch. In accordance with~\cite{hubert-2023-loss}, KGEMs are trained during $400$ epochs. For TransE, DistMult, and ComplEx, uniform random negative sampling~\cite{transe} was used to pair each train triple with one negative counterpart. ConvE and TuckER are trained using 1-N scoring~\cite{conve}. In this work, we stick with the loss function originally used for each KGEM. Evaluation is performed every $10$ epochs and the best model w.r.t. MRR on the validation set is saved for performance assessment on the test set. 
The aforementioned procedure prevails for the vanilla training mode. For P1 and P2, we use the same procedure as for training on the KG, except that $200$ epochs of training are previously performed on the protograph, for a total of $600$ training epochs.

\sstitle{Experimental results.} LP results are reported in Table~\ref{tab:LP-results}.
On YAGO14k and DBpedia77k, learning embeddings with P1 or P2 provides better Sem@$K$ values compared to V. No significant change is noticed on FB15k187. This is likely to be an artefact of the Freebase class hierarchy and how entities are typed in FB15k187: the class hierarchy only has 2 levels, all relation domains and ranges are level-2 classes (therefore P1 and P2 are the same, see Table~\ref{tab:datasets}) and entities share many classes together. In this situation, the protograph may not contain discriminative enough information for pre-training embeddings, which is reflected in the similar results achieved regardless of the training setting. In addition, we observe that some KGEMs seem to benefit more from the MASCHInE approach. In particular, ConvE and TuckER tend to perform better w.r.t. both rank-based and semantic-oriented metrics. Overall, we notice that including protographs in the training phase helps making semantically valid predictions, as evidenced by the better Sem@$K$ under P1 and P2.
Protographs may help organize the embedding space in such a way that geometric distances between embeddings better reflect their semantic similarities.
Regarding rank-based metrics, we notice equivalent performance, which could be explained since LP relies less on semantics and more on relational patterns~\cite{bhardwaj-2021} compared to other tasks such as entity clustering (see below). 

\subsection{Entity Clustering}
\label{entity-clustering}
Jain \textit{et al.}~\cite{jain-2021} demonstrate that embeddings do not consistently capture the semantics of the KG (in particular, they only considered schema-agnostic approaches). In this section, we investigate whether embeddings learnt for the LP task under P1 and P2 can actually prove useful in EC.

\subsubsection{Experiments on the original datasets.}
We investigate whether the embeddings learnt for the LP task (see Section~\ref{LP}) under P1 and P2 allow a better class separability for the task of EC.

\sstitle{Datasets.} We reuse YAGO14k, FB15k187, and DBpedia77k as presented in Section~\ref{LP}. The ground-truth labels are the most-generic classes an entity belongs to. It should be noted that entities belonging to several most-generic classes are filtered out as they would be assigned multiple ground-truth labels.

\sstitle{Implementation details.} Entity embeddings are retrieved at the best epoch on the validation sets of YAGO14k, FB15k187, and DBpedia77k, for all the KGEMs and under the three settings V, P1, and P2.
Then, following the evaluation protocol in Jain \textit{et al.}~\cite{jain-2021}, the k-means clustering algorithm is run using default parameters of scikit-learn\footnote{\url{https://scikit-learn.org/stable/index.html}}.

\begin{figure*}[t]
  \centering
  \begin{minipage}[b]{0.33\textwidth}
    \centering
    \includegraphics[width=0.7\textwidth]{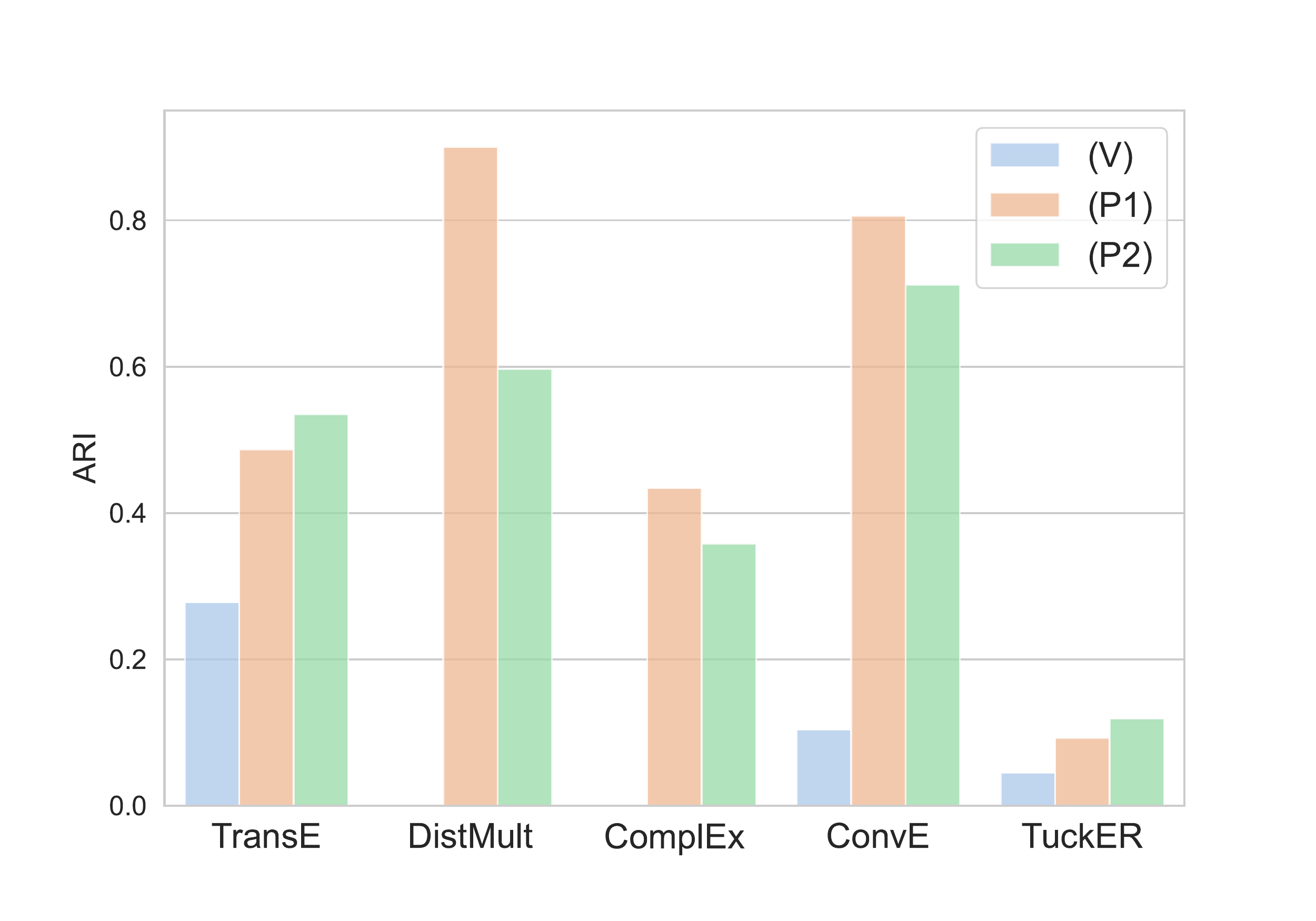}
    \vspace{-0.5cm}
    \caption*{\centering (a) YAGO14k}
    \label{subfig:YAGO14K-ari}
  \end{minipage}\hfill
  \begin{minipage}[b]{0.33\textwidth}
    \centering
    \includegraphics[width=0.7\textwidth]{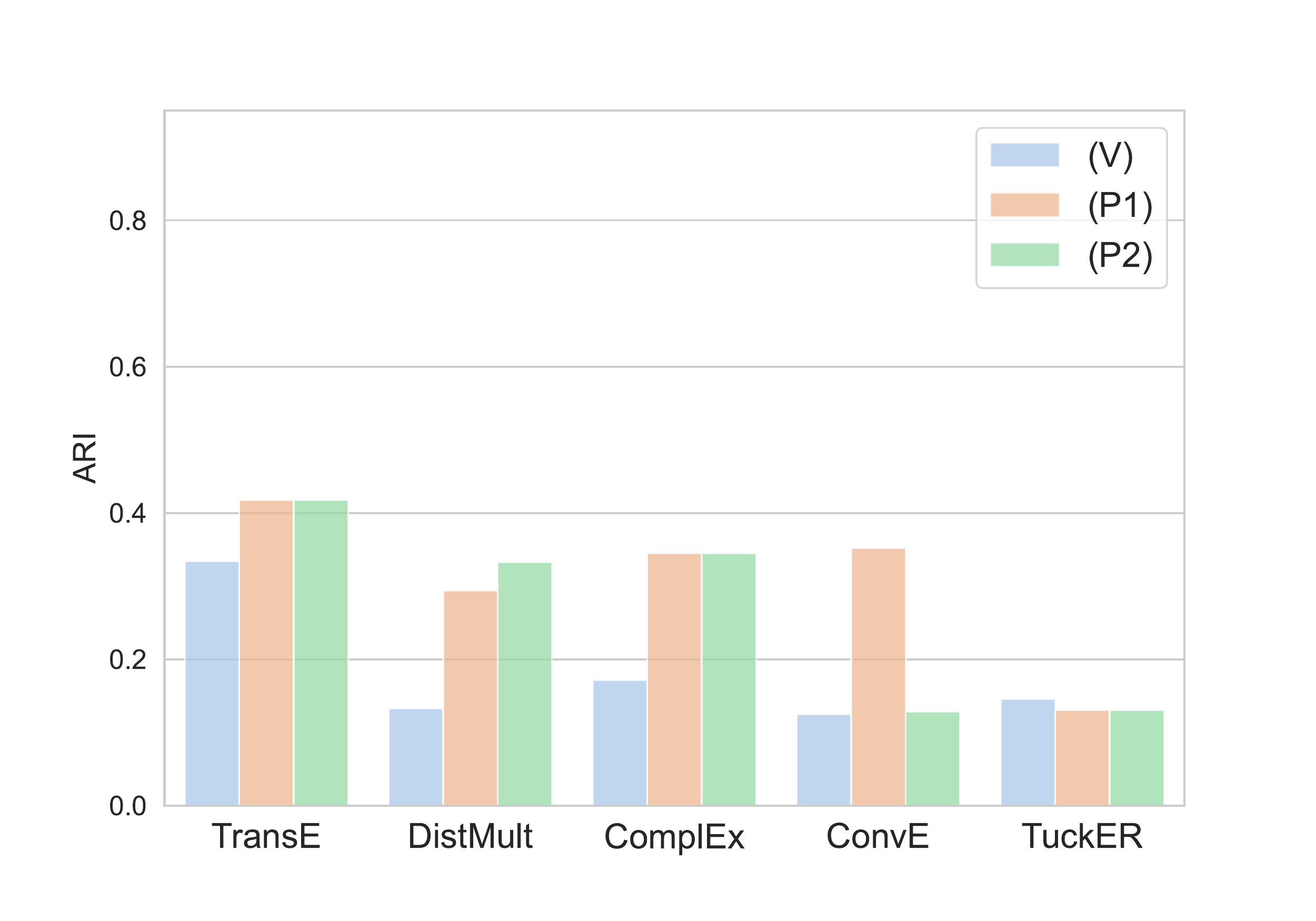}
    \vspace{-0.5cm}
    \caption*{\centering (b) FB15k187}
    \label{subfig:FB15k187-ari}
  \end{minipage}\hfill
  \begin{minipage}[b]{0.33\textwidth}
    \centering
    \includegraphics[width=0.7\textwidth]{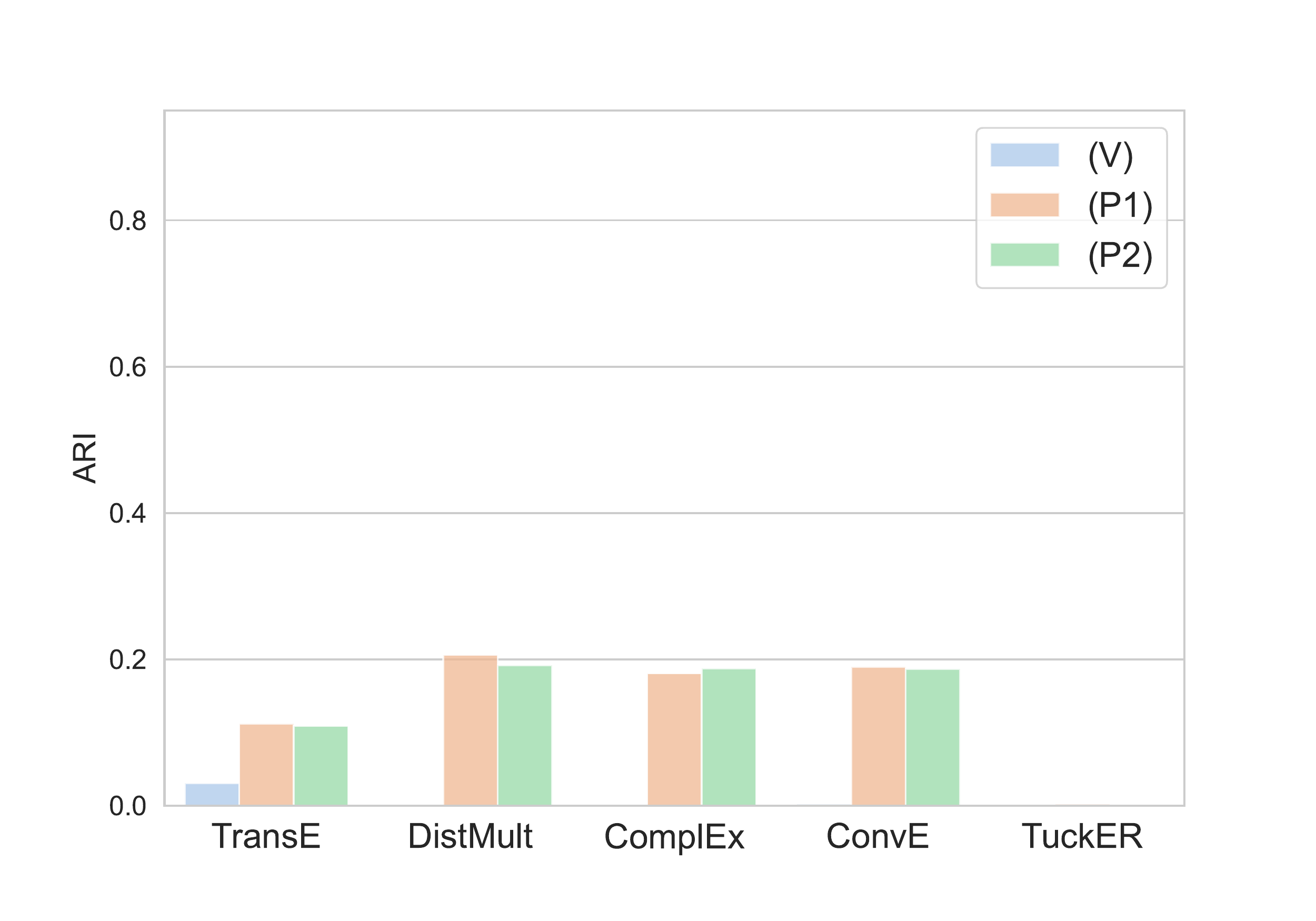}
    \vspace{-0.5cm}
    \caption*{\centering (c) DBpedia77k}
    \label{subfig:DBpedia77k-ari}
  \end{minipage}

  \vspace{-0.05cm}

  \begin{minipage}[b]{0.33\textwidth}
    \centering
    \includegraphics[width=0.7\textwidth]{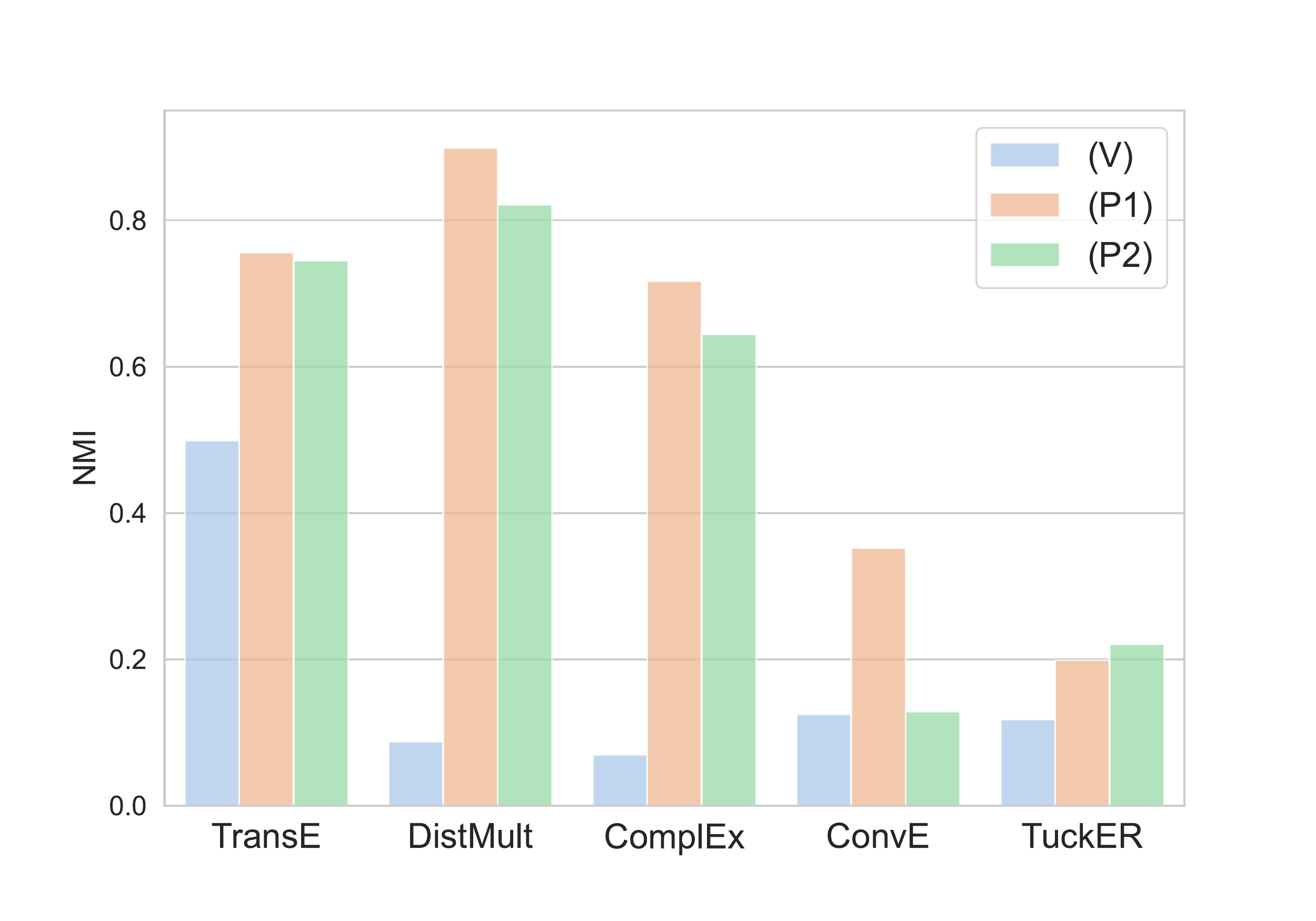}
    \vspace{-0.5cm}
    \caption*{\centering (d) YAGO14k}
    \label{subfig:YAGO14K-nmi}
  \end{minipage}\hfill
  \begin{minipage}[b]{0.33\textwidth}
    \centering
    \includegraphics[width=0.7\textwidth]{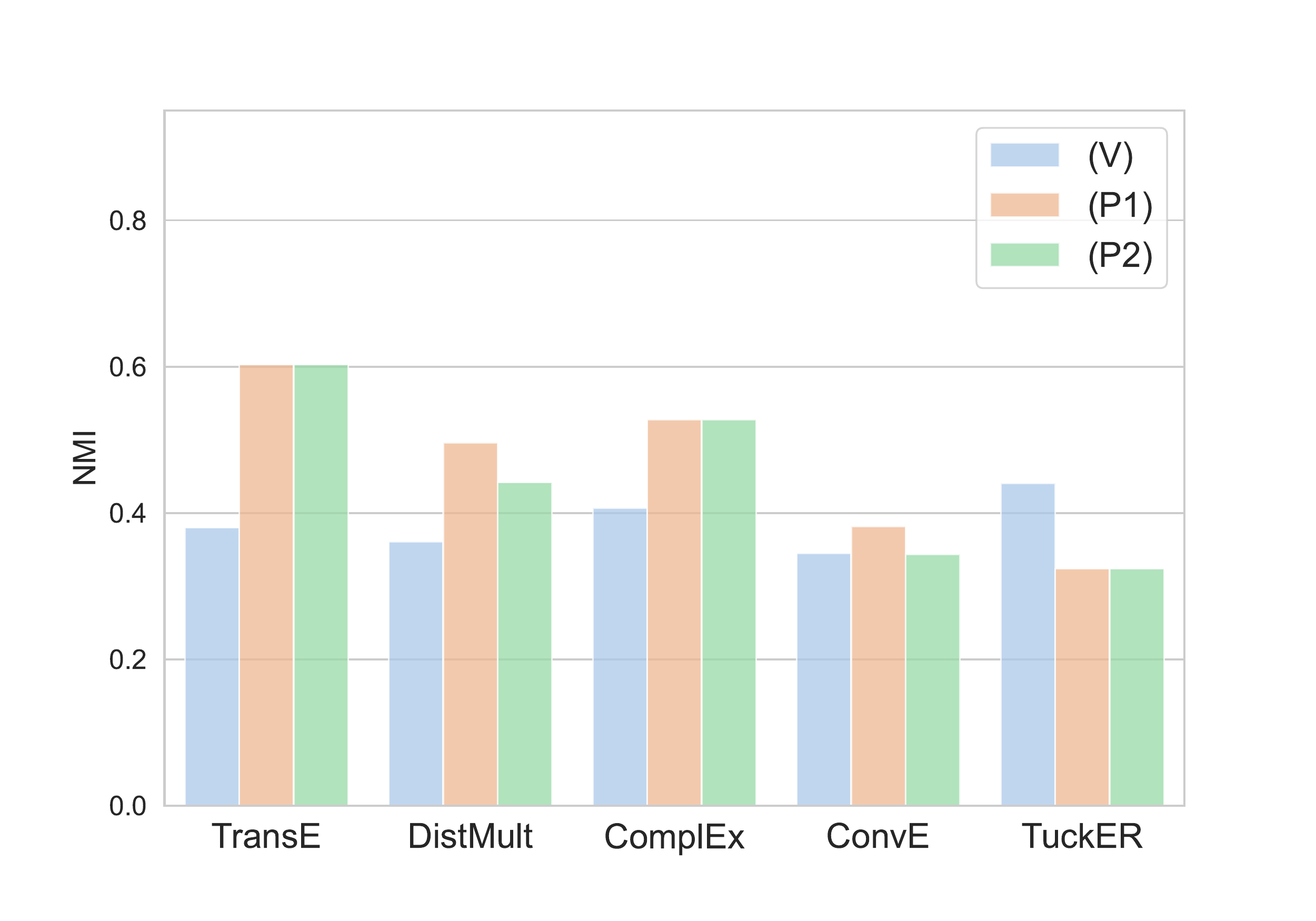}
    \vspace{-0.5cm}
    \caption*{\centering (e) FB15k187}
    \label{subfig:FB15k187-nmi}
  \end{minipage}\hfill
  \begin{minipage}[b]{0.33\textwidth}
    \centering
    \includegraphics[width=0.7\textwidth]{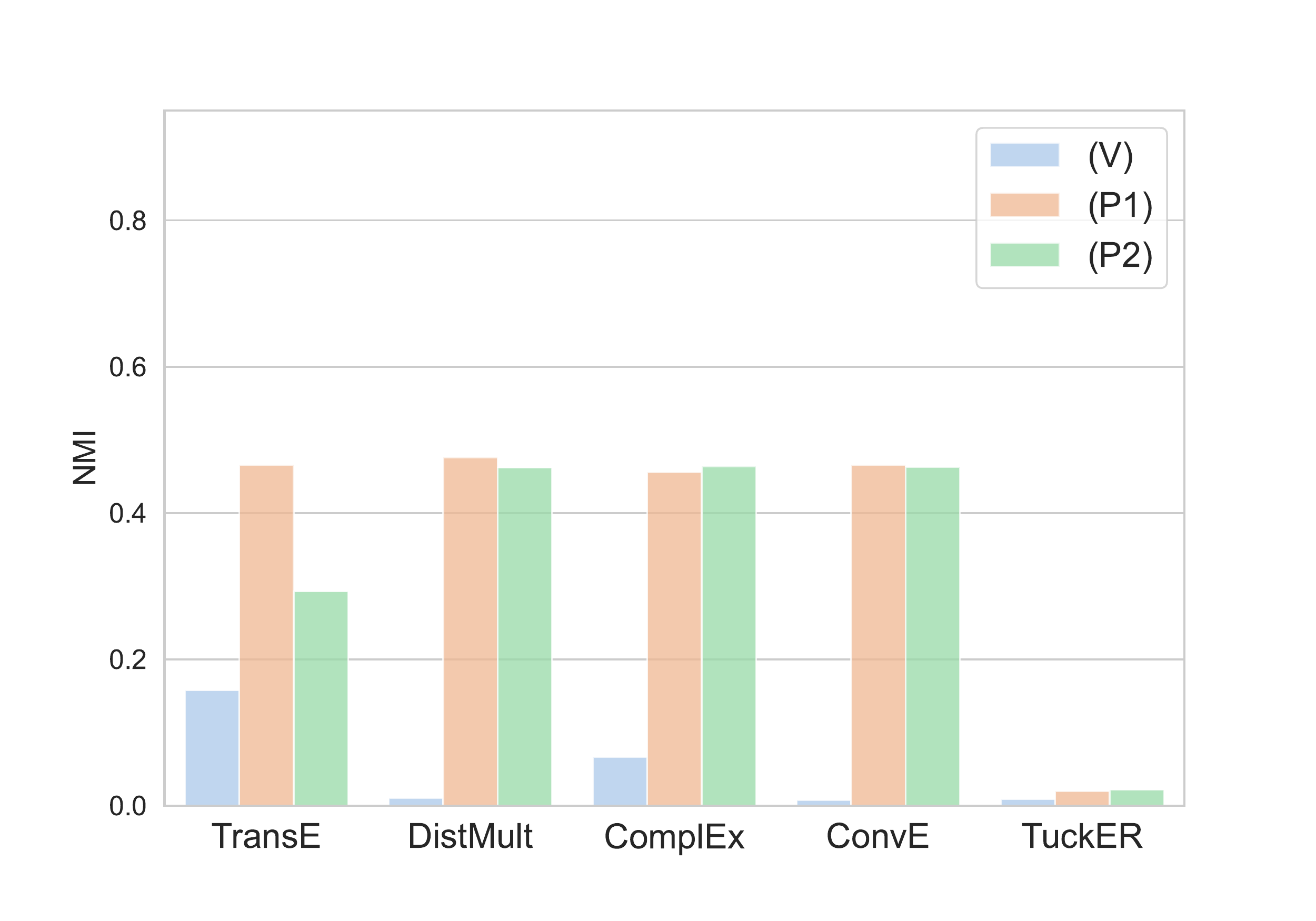}
    \vspace{-0.5cm}
    \caption*{\centering (f) DBpedia77k}
    \label{subfig:DBpedia77k-nmi}
  \end{minipage}
  \caption{Entity clustering results on YAGO14k, FB15k187, and DBpedia77k.}
  \label{fig:ari-nmi}
\end{figure*}

\sstitle{Evaluation metrics.} Clustering results are evaluated using the Adjusted Rand Index (ARI) -- which measures the similarity between the cluster assignments by making pairwise comparisons -- and Normalized Mutual Information (NMI) -- which measures the agreement between the cluster assignments. It should be noted that these metrics are applicable as we have the ground-truth labels for each entity. In particular, we compare clusters output by k-means with ground-truth labels that are the classes entities belong to. ARI and NMI results are reported in Figure~\ref{fig:ari-nmi}. Lowest and highest scores are $0$ ($-1$ for ARI) and $1$, respectively.

\begin{figure}[h]
  \centering
  \begin{minipage}[b]{0.21\textwidth}
    \centering
    \includegraphics[trim={4cm 2cm 3.5cm 3cm},clip, width=\textwidth]{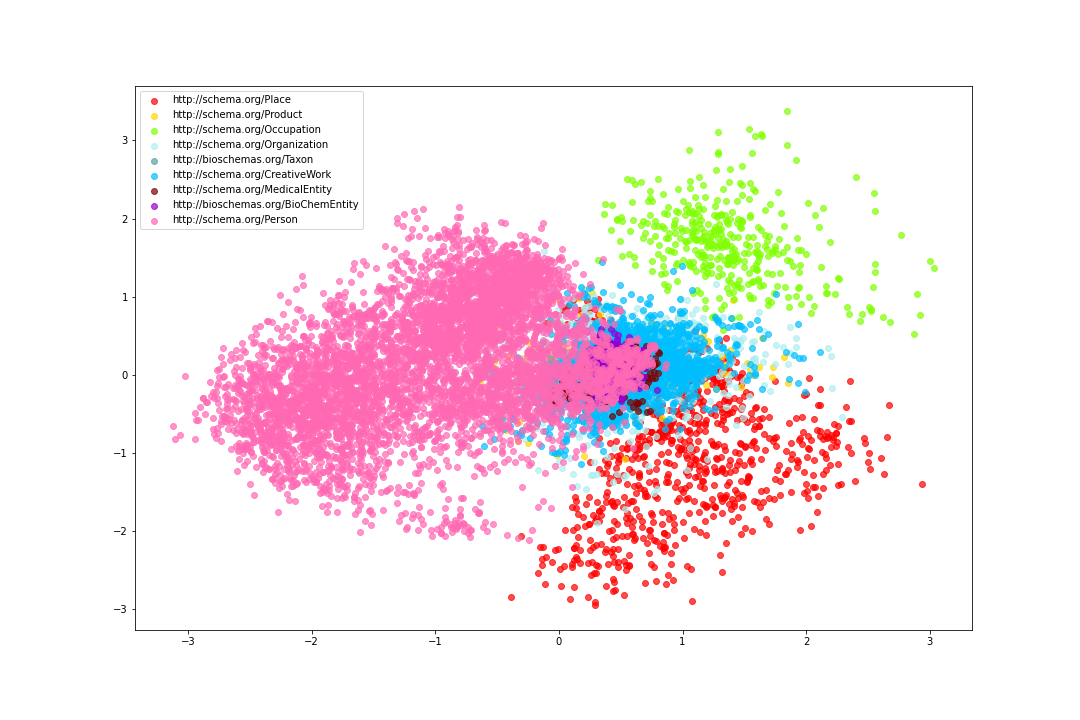}
    \vspace{-0.75cm}
    \caption*{\centering (a) TransE (V)}
    \label{subfig:transe-vanilla}
  \end{minipage}
  \qquad
  \begin{minipage}[b]{0.21\textwidth}
    \centering
    \includegraphics[trim={4cm 2cm 3.5cm 3cm},clip,width=\textwidth]{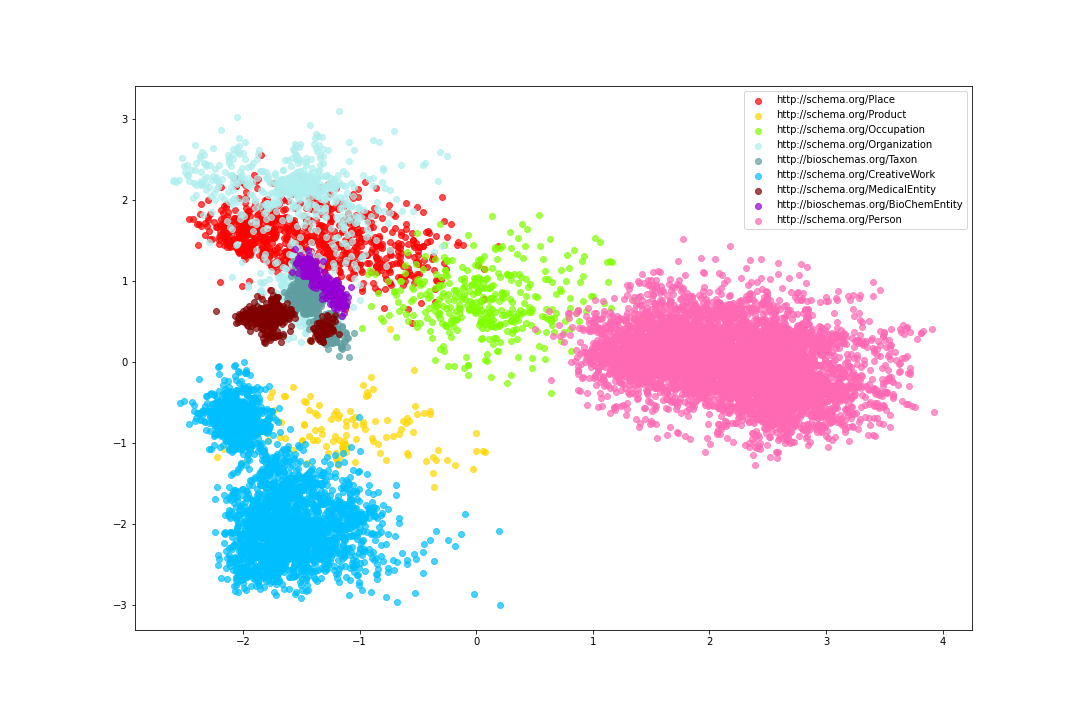}
    \vspace{-0.75cm}
    \caption*{\centering (b) TransE (P2)}
    \label{subfig:transe-prototype}
  \end{minipage}

    \vspace{0.3cm}

  \begin{minipage}[b]{0.21\textwidth}
    \centering
    \includegraphics[trim={4cm 2cm 3.5cm 3cm},clip,width=\textwidth]{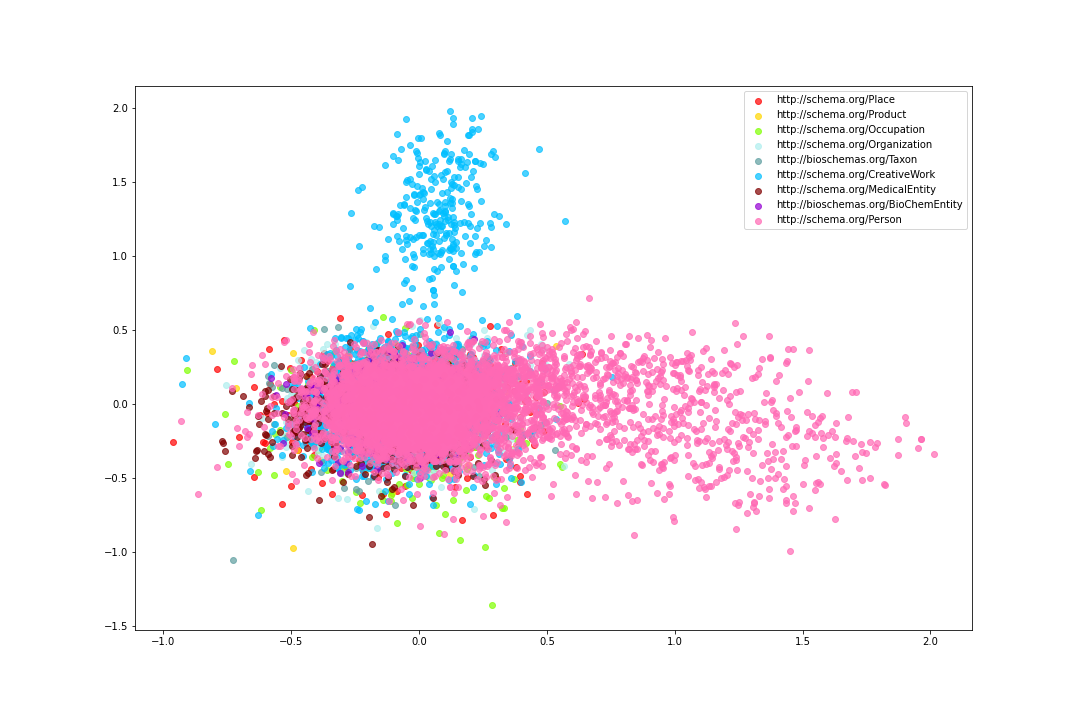}
    \vspace{-0.75cm}
    \caption*{\centering (c) DistMult (V)}
    \label{subfig:distmult-vanilla}
  \end{minipage}
  \qquad
  \begin{minipage}[b]{0.21\textwidth}
    \centering
    \includegraphics[trim={4cm 2cm 3.5cm 3cm},clip,width=\textwidth]{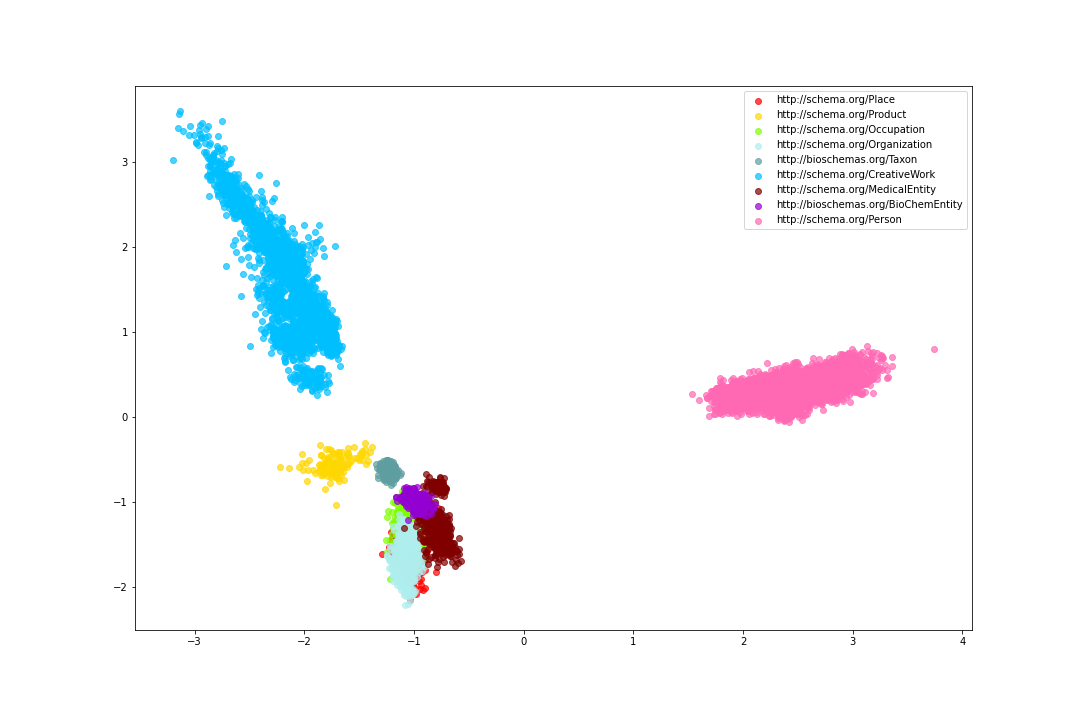}
    \vspace{-0.75cm}
    \caption*{\centering (d) DistMult (P2)}
    \label{subfig:distmult-prototype}
  \end{minipage}
  \vspace{-0.1cm}
  \caption{PCA visualizations of YAGO14k entity embeddings using the first two principal components.
  Each point represents an entity, whose color denotes its ground-truth class.}
  \label{fig:pca}
\end{figure}

\sstitle{Experimental results.} Except for TuckER on FB15k187, both settings P1 and P2 provide substantial improvements. It seems that the entity embeddings learnt by TuckER are not able to cluster entities based on their classes, which diminishes the relevancy of using this model for drawing comparisons. That being said, the benefit of P1 and P2 is consistent across datasets. This is even more striking for YAGO14k, as evidenced in Figure~\ref{fig:pca}: entities sharing the same ground-truth class tend to cluster more under P2. Jain \textit{et al.}~\cite{jain-2021} pointed out the limitations of using KGEs for semantic representation of the underlying entities and relations. The relatively low performance of embeddings for EC under V (Figure~\ref{fig:ari-nmi} and Figure~\ref{fig:pca}) indeed suggests that the semantic similarity between entities is not reflected through geometric similarities in their embeddings. However, P1 and P2 generate entity latent representations whose geometric similarities are more representative of their classes.

\begin{table*}[t]
\centering
\caption{Entity clustering results on the GEval datasets.
Bold font indicates which setting performs the best for a given model and dataset, while underlined results correspond to the second best setting.}\label{tab:geval-results}
\footnotesize
\setlength{\tabcolsep}{2.5pt}
\begin{tabular}{ccccccccccccccccccccccc}
\toprule
& & & \multicolumn{6}{c}{CC} && \multicolumn{6}{c}{CCB} && \multicolumn{6}{c}{CAMAP} \\
\cmidrule(lr){4-9} \cmidrule(lr){11-16} \cmidrule(lr){18-23}
& & & ARI & AMI & VM & FM & H & C & & ARI & AMI & VM & FM & H & C & & ARI & AMI & VM & FM & H & C\\
\midrule
\multirow{3}{*}{TransE} & V && $0.179$ & $0.140$ & $0.141$ & $0.596$ & $0.143$ & $0.139$ & & $0.183$ & $0.147$ & $0.147$ & $0.649$ & $0.164$ & $0.133$ && $0.564$ & $0.783$ & $0.785$ & $0.736$ & $0.951$ & $0.668$\\
 & P1 && $\underline{0.621}$ & $\underline{0.514}$ & $\underline{0.515}$ & $\underline{0.823}$ & $\underline{0.505}$ & $\underline{0.526}$ & & $\underline{0.288}$ & $\underline{0.311}$ & $\underline{0.311}$ & $\underline{0.692}$ & $\underline{0.349}$ & $\underline{0.280}$ && $\underline{0.593}$ & $\underline{0.795}$ & $\underline{0.796}$ & $\underline{0.756}$ & $\underline{0.956}$ & $\underline{0.682}$\\
 & P2 && $\mathbf{0.680}$ & $\mathbf{0.570}$ & $\mathbf{0.571}$ & $\mathbf{0.849}$ & $\mathbf{0.563}$ & $\mathbf{0.579}$ & & $\mathbf{0.299}$ & $\mathbf{0.314}$ & $\mathbf{0.314}$ & $\mathbf{0.698}$ & $\mathbf{0.353}$ & $\mathbf{0.284}$ && $\mathbf{0.646}$ & $\mathbf{0.815}$ & $\mathbf{0.816}$ & $\mathbf{0.791}$ & $\mathbf{0.964}$ & $\mathbf{0.708}$\\
\midrule
\multirow{3}{*}{DistMult} & V && $0.018$ & $0.024$ & $0.027$ & $0.709$ & $0.015$ & $0.095$ && $-0.024$ & $0.009$ & $0.010$ & $0.770$ & $0.006$ & $0.028$ && $-0.054$ & $0.099$ & $0.106$ & $0.478$ & $0.087$ & $0.136$\\
 & P1 && $\mathbf{0.891}$ & $\underline{0.817}$ & $\underline{0.817}$ & $\mathbf{0.948}$ & $\underline{0.813}$ & $\underline{0.821}$ && $\mathbf{0.762}$ & $\mathbf{0.673}$ & $\mathbf{0.673}$ & $\mathbf{0.907}$ & $\mathbf{0.708}$ & $\mathbf{0.642}$ && $\underline{0.655}$ & $\underline{0.705}$ & $\underline{0.707}$ & $\underline{0.798}$ & $\underline{0.782}$ & $\underline{0.646}$\\
 & P2 && $\mathbf{0.891}$ & $\mathbf{0.824}$ & $\mathbf{0.824}$ & $\mathbf{0.948}$ & $\mathbf{0.818}$ & $\mathbf{0.830}$ && $\underline{0.756}$ & $\underline{0.626}$ & $\underline{0.627}$ & $\mathbf{0.907}$ & $\underline{0.646}$ & $\underline{0.608}$ && $\mathbf{0.933}$ & $\mathbf{0.872}$ & $\mathbf{0.872}$ & $\mathbf{0.963}$ & $\mathbf{0.886}$ & $\mathbf{0.859}$\\
\midrule
\multirow{3}{*}{ComplEx} & V && $-0.011$ & $0.009$ & $0.011$ & $0.694$ & $0.007$ & $0.033$ && $-0.024$ & $0.009$ & $0.010$ & $0.771$ & $0.006$ & $0.028$ && $0.139$ & $0.345$ & $0.350$ & $0.520$ & $0.341$ & $0.360$\\
 & P1 && $\mathbf{0.897}$ & $\mathbf{0.827}$ & $\mathbf{0.828}$ & $\mathbf{0.951}$ & $\mathbf{0.823}$ & $\mathbf{0.832}$ && $\mathbf{0.619}$ & $\mathbf{0.560}$ & $\mathbf{0.560}$ & $\mathbf{0.845}$ & $\mathbf{0.606}$ & $\mathbf{0.521}$ && $\mathbf{0.969}$ & $\mathbf{0.916}$ & $\mathbf{0.917}$ & $\mathbf{0.983}$ & $\mathbf{0.939}$ & $\mathbf{0.896}$\\
 & P2 && $\underline{0.319}$ & $\underline{0.264}$ & $\underline{0.265}$ & $\underline{0.665}$ & $\underline{0.269}$ & $\underline{0.262}$ && $\underline{0.493}$ & $\underline{0.460}$ & $\underline{0.460}$ & $\underline{0.788}$ & $\underline{0.507}$ & $\underline{0.422}$ && $\underline{0.961}$ & $\underline{0.914}$ & $\underline{0.914}$ & $\underline{0.978}$ & $\underline{0.938}$ & $\underline{0.892}$\\
\midrule
\multirow{3}{*}{ConvE} & V && $0.311$ & $0.234$ & $0.235$ & $0.691$ & $0.223$ & $0.249$ && $0.271$ & $0.242$ & $0.243$ & $0.687$ & $0.271$ & $0.220$ && $0.591$ & $0.608$ & $0.611$ & $0.76$ & $0.672$ & $0.559$\\
 & P1 && $\underline{0.878}$ & $\underline{0.805}$ & $\underline{0.805}$ & $\underline{0.942}$ & $\underline{0.799}$ & $\underline{0.811}$ && $\underline{0.719}$ & $\underline{0.628}$ & $\underline{0.628}$ & $\underline{0.889}$ & $\underline{0.665}$ & $\underline{0.596}$ && $\underline{0.934}$ & $\underline{0.880}$ & $\underline{0.881}$ & $\underline{0.964}$ & $\underline{0.928}$ & $\underline{0.839}$\\
 & P2 && $\mathbf{0.916}$ & $\mathbf{0.848}$ & $\mathbf{0.848}$ & $\mathbf{0.960}$ & $\mathbf{0.847}$ & $\mathbf{0.849}$ && $\mathbf{0.790}$ & $\mathbf{0.697}$ & $\mathbf{0.697}$ & $\mathbf{0.919}$ & $\mathbf{0.727}$ & $\mathbf{0.669}$ && $\mathbf{0.954}$ & $\mathbf{0.915}$ & $\mathbf{0.915}$ & $\mathbf{0.975}$ & $\mathbf{0.968}$ & $\mathbf{0.868}$\\
\midrule
\multirow{3}{*}{TuckER} & V && $\mathbf{0.179}$ & $\mathbf{0.138}$ & $\mathbf{0.139}$ & $\mathbf{0.596}$ & $\mathbf{0.141}$ & $\mathbf{0.137}$ && $\mathbf{0.158}$ & $\mathbf{0.120}$ & $\mathbf{0.121}$ & $\mathbf{0.639}$ & $\mathbf{0.135}$ & $\mathbf{0.109}$ && $\mathbf{0.396}$ & $0.526$ & $0.529$ & $\mathbf{0.618}$ & $0.636$ & $0.453$\\
 & P1 && $\underline{0.069}$ & $\underline{0.051}$ & $\underline{0.052}$ & $\underline{0.543}$ & $\underline{0.053}$ & $\underline{0.051}$ && $\underline{0.137}$ & $\underline{0.094}$ & $\underline{0.094}$ & $\underline{0.631}$ & $\underline{0.105}$ & $\underline{0.086}$ && $\underline{0.376}$ & $\mathbf{0.635}$ & $\mathbf{0.637}$ & $\underline{0.599}$ & $\mathbf{0.795}$ & $\mathbf{0.531}$\\
 & P2 && $0.047$ & $0.042$ & $0.043$ & $0.533$ & $0.043$ & $0.042$ && $0.033$ & $0.051$ & $0.052$ & $0.578$ & $0.058$ & $0.046$ && $0.276$ & $0.487$ & $0.490$ & $0.520$ & $0.621$ & $0.405$\\
\bottomrule
\end{tabular}
\end{table*}

\subsubsection{Experiments on GEval clustering problems}
In this follow-up entity clustering experiment, the GEval evaluation framework~\cite{pellegrino-2020} is used. GEval provides gold standard datasets based on DBpedia and suggest clustering models, configurations and evaluation metrics\footnote{\url{https://github.com/mariaangelapellegrino/Evaluation-Framework}}.

\sstitle{Datasets.} GEval proposes 4 datasets based on DBpedia. The first dataset contain entities of classes \texttt{dbo:AmericanFootballTeam} and \texttt{dbo:BasketballTeam}. We do not use this dataset as DBpedia77k does not contain any of these entities. Instead, we use the other three datasets for which DBpedia77k has a biggest coverage. For clarity, we name these datasets as CC, CCB, and CAMAP. The first two ones contain different numbers of DBpedia entities of classes \texttt{dbo:Cities} and \texttt{dbo:Countries}, while the third dataset contains five types of entities. Therefore, there are two ground-truth clusters in the first two datasets and five clusters in the third one.

\sstitle{Implementation details.} We rely on GEval guidelines and perform the preliminary filtering step described as follows. First, we select entity embeddings learnt on DBpedia77k for the LP task (see Section~\ref{LP}).
Embeddings of entities that do not appear in the GEval benchmark datasets are filtered out. Clustering with all the algorithms suggested in the GEval framework is performed on the remaining entities and evaluated w.r.t. several metrics. 

\sstitle{Evaluation metrics.} As in the previous experiment, we use ARI. Following the GEval guidelines, we additionally use Adjusted Mutual Information Score (AMI), V-Measure (VM), Fowlkes-Mallow Score (FM), Homogeneity (H), and Completeness (C). V-Measure is the harmonic mean of homogeneity and completeness measure, while the Fowlkes-Mallow Score is the geometric mean of pairwise precision and recall. All these metrics measure the correctness of the cluster assignments based on ground-truth labels. For each of these datasets, and under each setting, results achieved with the best performing clustering method are shown in Table~\ref{tab:geval-results}. Lowest and highest scores are $0$ ($-1$ for ARI) and $1$, respectively.

\begin{table*}[h]
\centering
\caption{Node classification results on the 12 DLCC test cases. Listed are the results of the best classifier for each combination of model and test case. Bold fonts indicate which setting performs the best for a given model and test case. Underlined results show which combination of model and setting performs the best for each test case.}\label{tab:dlcc-results}
\footnotesize
\begin{tabular}{cccccccccccccc}
\toprule
Test Case & Setting & TransE & DistMult & ComplEx & ConvE & TuckER & Test Case & Setting & TransE & DistMult & ComplEx & ConvE & TuckER \\
\midrule
\multirow{3}{*}{tc01} & V & $\mathbf{0.702}$ & $\mathbf{0.810}$ & $\mathbf{0.820}$ & $\mathbf{\underline{0.885}}$ & $0.832$ & \multirow{3}{*}{tc02} & V & $\mathbf{0.654}$ & $0.564$ & $0.566$ & $0.732$ & $0.669$ \\ 
 & P1 & $0.647$ & $0.754$ & $0.784$ & $0.875$ & $\mathbf{0.857}$ & & P1 & $0.566$ & $0.554$ & $0.554$ & $\mathbf{\underline{0.764}}$ & $0.637$ \\
 & P2 & $0.637$ & $0.794$ & $0.769$ & $0.852$ & $0.792$ & & P2 & $0.571$ & $\mathbf{0.627}$ & $\mathbf{0.627}$ & $0.687$ & $\mathbf{0.679}$ \\
\midrule
\multirow{3}{*}{tc03} & V & $0.584$ & $\mathbf{0.571}$ & $\mathbf{0.554}$ & $0.629$ & $\mathbf{0.669}$ & \multirow{3}{*}{tc04} & V & $0.638$ & $0.540$ & $0.512$ & $0.818$ & $0.570$ \\
 & P1 & $\mathbf{0.589}$ & $0.541$ & $0.519$ & $\mathbf{\underline{0.699}}$ & $0.664$ & & P1 & $0.760$ & $\mathbf{0.812}$ & $\mathbf{0.832}$ & $\mathbf{\underline{0.870}}$ & $0.552$ \\
 & P2 & $0.521$ & $0.544$ & $0.526$ & $0.622$ & $0.556$ & & P2 & $\mathbf{0.805}$ & $0.802$ & $0.818$ & $0.818$ & $\mathbf{0.680}$ \\
\midrule
\multirow{3}{*}{tc05} & V & $0.704$ & $0.683$ & $0.678$ & $0.822$ & $0.814$ & \multirow{3}{*}{tc06} & V & $0.645$ & $\mathbf{0.940}$ & $\mathbf{0.938}$ & $\mathbf{0.940}$ & $\mathbf{\underline{0.950}}$ \\
 & P1 & $0.887$ & $0.937$ & $\mathbf{\underline{0.940}}$ & $\mathbf{\underline{0.940}}$ & $0.814$ & & P1 & $0.718$ & $0.875$ & $0.898$ & $0.935$ & $\mathbf{\underline{0.950}}$ \\
 & P2 & $\mathbf{0.937}$ & $\mathbf{\underline{0.940}}$ & $\mathbf{\underline{0.940}}$ & $\mathbf{\underline{0.940}}$ & $\mathbf{0.902}$ & & P2 & $\mathbf{0.740}$ & $0.828$ & $0.788$ & $0.895$ & $0.942$ \\
\midrule
\multirow{3}{*}{tc07} & V & $0.578$ & $0.515$ & $0.552$ & $0.560$ & $0.522$ & \multirow{3}{*}{tc08} & V & $0.552$ & $0.552$ & $0.520$ & $\mathbf{0.598}$ & $0.575$ \\
 & P1 & $0.505$ & $0.540$ & $0.560$ & $0.560$ & $0.552$ & & P1 & $0.585$ & $0.560$ & $0.558$ & $0.575$ & $0.580$ \\
 & P2 & $\mathbf{0.582}$ & $\mathbf{0.605}$ & $\mathbf{0.655}$ & $\mathbf{\underline{0.675}}$ & $\mathbf{0.648}$ & & P2 & $\mathbf{0.595}$ & $\mathbf{0.588}$ & $\mathbf{0.588}$ & $0.578$ & $\mathbf{\underline{0.672}}$ \\
\midrule
\multirow{3}{*}{tc09} & V & $0.552$ & $0.522$ & $0.525$ & $0.730$ & $0.745$ & \multirow{3}{*}{tc10} & V & $0.578$ & $\mathbf{0.572}$ & $\mathbf{0.562}$ & $\mathbf{0.640}$ & $0.690$ \\
 & P1 & $\mathbf{0.560}$ & $0.538$ & $0.552$ & $\mathbf{0.745}$ & $\mathbf{\underline{0.780}}$ & & P1 & $\mathbf{0.598}$ & $0.550$ & $0.538$ & $0.625$ & $\mathbf{\underline{0.692}}$ \\
 & P2 & $0.550$ & $\mathbf{0.615}$ & $\mathbf{0.598}$ & $0.672$ & $0.742$ & & P2 & $0.538$ & $0.510$ & $0.535$ & $0.590$ & $0.588$ \\
\midrule
\multirow{3}{*}{tc11} & V & $\mathbf{0.632}$ & $0.518$ & $0.535$ & $\mathbf{0.688}$ & $\mathbf{\underline{0.700}}$ & \multirow{3}{*}{tc12} & V & $\mathbf{0.650}$ & $0.538$ & $0.552$ & $0.702$ & $0.708$ \\
 & P1 & $0.628$ & $0.542$ & $0.522$ & $0.618$ & $0.685$ & & P1 & $0.572$ & $0.530$ & $0.565$ & $0.708$ & $0.700$ \\
 & P2 & $0.630$ & $\mathbf{0.552}$ & $\mathbf{0.558}$ & $0.638$ & $0.678$ &  & P2 & $0.625$ & $\mathbf{\underline{0.802}}$ & $\mathbf{0.738}$ & $\mathbf{\underline{0.802}}$ & $\mathbf{0.752}$ \\
\bottomrule
\end{tabular}
\end{table*}
 
\sstitle{Experimental results.} 
First, it is worth noting that results on CAMAP are significantly higher than on the other two datasets. Recall that CAMAP includes entities belonging to five different classes, whereas CC and CCB only contain entities from \texttt{dbo:Cities} and \texttt{dbo:Countries}. In this experiment, the embeddings learnt by the KGEMs are most useful for performing EC with five ground-truth clusters than with two, which might seem counterintuitive. However, in~\cite{jain-2021}, it is demonstrated that in a KG, only a small subset of its entities actually have an embedding with a meaningful semantic representation. This is the reason why for the clustering task, results are not consistent across subsets of entities~\cite{jain-2021}. This is in line with our experimental findings that lead to the same conclusion. 

\cite{zouaq-2020} is also concerned with the ability of KGEMs to embed entities from the same class to the same region of the embedding space. Their results suggest that entities belonging to rare classes have embeddings that are less separable w.r.t. to entity embeddings of other classes compared with entities belonging to frequent classes. This assumption needs to be qualified in light of our experimental findings: lots of entities in DBpedia77k are \texttt{dbo:City} and \texttt{dbo:Country}. Following \cite{zouaq-2020}, we should expect better results for CC and CCB in Table~\ref{tab:geval-results}. However, it should be noted that both \texttt{dbo:City} and \texttt{dbo:Country} are subclasses of \texttt{dbo:Place} (equivalent to \texttt{dbo:Location}), which is over-represented in DBpedia77k\footnote{In DBpedia77k, $11,586$ entities belong to the \texttt{dbo:Place} class.}. It is arguably harder to separate entity embeddings for CC and CCB, as all of them actually belong to the same parent class. In contrast, CAMAP contain albums, cities, companies, movies, and universities. In this case, \texttt{dbo:City} is a child class of \texttt{dbo:Place}, \texttt{dbo:Company} and \texttt{dbo:University} are child classes of \texttt{dbo:Agent} while \texttt{dbo:Album} and \texttt{dbo:Movie} are child classes of \texttt{dbo:Work}. With three different generic classes, the semantic similarities between entities in CAMAP is more easily reflected into the geometric similarities of their embeddings. 

Interestingly, the discrepancies in results, \textit{i.e.}, better results on CAMAP compared with CC and CCB, are setting-dependent (V, P1, P2). Indeed, they are alleviated when using either P1 or P2. Under V, TransE, ComplEx, and ConvE provide substantially better results on CAMAP compared to CC. Once using P2, we cannot see any significant difference in results on CAMAP compared to CC for these models. This means that training with a protograph-assisted approach led the KGEMs to generate higher-quality embeddings in terms of class separability, especially for entities in CC and CCB that are harder to distinguish.
With the sole exception of TuckER whose performance may be due to non-optimal hyperparameter tuning, strong improvements are achieved w.r.t. the EC task on all datasets. Interestingly, not only V is outperformed by either P1 or P2 in each scenario, but actually by both of them at the same time. As a result, our schema-based, protograph-assisted learning approach MASCHInE proves effective for EC, regardless of the protograph design principle.

\subsection{Node Classification}
Following on from the previous clustering experiment that clearly showed the benefit of incorporating pre-trained protograph embeddings, in this section we are interested in performing node classification with such embeddings. In this experiment, entities are labelled on the basis of whether they comply with some description logic (DL) constructors.

\sstitle{Datasets.} In this experiment, we use the synthetic DLCC node classification benchmark~\cite{portisch-2022}. The DLCC benchmark helps analyze the quality of KGEs by evaluating them w.r.t. which kinds of DL constructors they can represent. We use the synthetic gold standard datasets\footnote{\url{https://github.com/janothan/DL-TC-Generator}}. These are 12 synthetic datasets featuring labelled entities to classify and underpinned by a schema defining entity types, relation domains and ranges. The synthetic part of DLCC has been chosen because the authors argue that those pose classification problems which are not influenced by any side effects and correlations in the knowledge graph.

\sstitle{Implementation details.} For these 12 test cases, their respective protographs are built. Then, the KGEMs are trained w.r.t. the LP task under the three different settings V, P1, and P2. In line with~\cite{portisch-2022}, multiple classifiers were trained for each test case: decision trees (DT), naive bayes (NB), k-nearest neighbors (k-NN), support vector machine (SVM), and multilayer perceptron (MLP). Based on the best entity embeddings learnt during LP, these models are used to classify entities using the default scikit-learn parameters. 

\sstitle{Evaluation metrics.} NC performance is measured using F-score, which is the harmonic mean of precision and recall. Lowest and highest scores are $0$ and $1$, respectively. 
For each model and test case, results achieved with the best classifier are reported in Table~\ref{tab:dlcc-results}.

\sstitle{Experimental results.} First, not all of the gold standard test cases are equally hard for the task of NC. For instance, entities in tc01, tc05, and tc06 are obviously easier to label than in the other datasets. Particularly, the DLCC synthetic gold standard datasets are built on the basis of different DL constructors in order to analyze which KG constructs are more easily recognized by KGEMs. 

For the less trivial cases, the gains that can be achieved by using protographs are more significant. Especially in cases which are fairly badly solved by vanilla embeddings (\textit{e.g.}, tc07, tc12), there are gains of more than 10 percentage points in F1. It is further remarkable that on this problem, compared to link prediction and entity clustering, TuckER can also benefit from the protographs.

When looking at which protograph construction strategy is more suitable for which kind of problem, we see that for those test cases which use a concrete class in their definition (particularly tc07, tc08, tc11, and tc12), P2 is clearly superior to P1. This may be due to the fact that P1 does not create class embeddings for all classes used in the respective constructors. Therefore, the resulting entity embeddings are less discriminating.

\section{Conclusion and Future Work}
\label{conclusion}
In this paper, we present MASCHInE -- a novel model-agnostic and protograph-assisted approach for learning KGEs. MASCHInE first trains embeddings on protographs and then transfers these embeddings to learn the final KGEs. We have shown that this approach produces more versatile KGEs, which can yield significant improvements in entity clustering and node classification. This is particularly noticeable for entity clustering, as protographs encode information about classes, on which entity clustering heavily relies on. We also demonstrated the advantage of using MASCHInE for link prediction: the use of protographs in the training procedure leads KGEMs to make more semantically valid predictions, which clearly highlights that geometric distance in the embedding space can be influenced by the semantics of the underlying KG.

In future work, we will extend MASCHInE by adding new protograph design principles. While in the current form, the transfer from the RDFS-based protograph embeddings to the KG entity embeddings is done once, we will study more expressive ontological constructs (\textit{e.g.}, OWL) and alternatives for KG and protograph iterative co-training. Additionally, it would be worth exploring whether pre-training over protographs provides benefits compared to passing the relevant RDFS triples to the KGEM as part of the whole graph. Comparing our approach with KGEMs that take advantage of additional information in the KG -- \textit{e.g.} attributes -- also deserves further exploration.

\bibliographystyle{ACM-Reference-Format}
\bibliography{sample-base}

\appendix

\end{document}